\documentclass{kluwer}    

\newdisplay{guess}{Conjecture}

\begin{document}                                                                                   
\begin{article}
\begin{opening}         
\title{A Massive Local Rules Search Approach to the Classification Problem}
\author{Vladislav G. \surname{Malyshkin}\footnote{ e-mail: mal@poly-aniline.com}} 
\author{Ray \surname{Bakhramov}}
\institute{Columbus Advisors LLC \\ 
	500 West Putnam ave \\
	Greenwich CT 06830}

\author{Andrey E. \surname{Gorodetsky}}
\institute{PME RAS St.Petersburg, Russia, 199178}

\runningauthor{Vladislav G. Malyshkin et al}
\runningtitle{A Massive Local Rules Search Approach to the Classification Problem}
\date{July 6, 2001}

\begin{abstract}
An approach to the classification problem of machine learning, based on
building local classification rules, is developed. The local rules are
considered as projections of the global classification rules to the event
we want to classify. A massive global optimization 
algorithm is used for optimization of quality criterion.
The algorithm, which has polynomial complexity in typical case,
is used to find all high--quality local rules. The other distinctive
feature of the algorithm is the integration of attributes levels
selection (for ordered attributes) with rules searching and original
conflicting rules resolution strategy. The algorithm is practical; it
was tested on a number of data sets from UCI repository, and a
comparison with the other predicting techniques is presented.
\end{abstract}
\end{opening}           

\section{Introduction}

Extraction of structural information from raw data
is a problem which is of great interest
for both fundamental and applied studies.
This paper will focus on one 
specific example of this problem --- classification.
The goal is to predict a class of a particular event.
This problem was approached from a number of 
different disciplines, including
Statistical Data Analysis \cite{dobson,ln},
Machine Learning \cite{carb,shav,aha_edt,mitch},
Fuzzy Logic \cite{fmi},
Operations Research \cite{or_2}
and Data Mining \cite{ps,fayyad}.
As a result, a variety
of learning techniques was developed.
The result of learning can be represented 
in a number of different forms.
The form that we are interested in working with is a set of rules.
It should be stressed that some other forms 
(such as decision trees, fuzzy models and many others)
are equivalent to a set of rules.

A set of rules (or any other form to which 
it is equivalent) is often
a preferred form of knowledge representation because it allows for 
a simple answer to the question, ``What was learned?''
 This specific set of rules was learned from the data.
For an algorithm, which produces only an answer,
it is often impossible to understand what was
really learned and why this specific answer was produced.
(The two mentioned knowledge representations differ as follows:
in the case that the result is a rule, the 
learned knowledge
is represented in a language which is richer
than one used to describe the dataset;
in the case that the result is a value, the
learned knowledge
is represented in the same language 
as the one used to describe the dataset \cite{ql_ib_rb}.)

The model--based techniques,  
such as developed in \cite{c45R8,segal}, 
take training data as input and 
produce a set of rules (or statements which 
are equivalent to rules) which can classify any event.
The lazy instance--based techniques, such as
developed in \cite{aha,aha_edt,ql_ib_rb},
return a result tailored to the specific event 
we want to classify. With such techniques 
the events similar to the given one are usually found first,
 then a prediction based on found instances is made.
An interesting attempt to combine
model based and lazy instance based learning 
was presented in \cite{melli}. In \cite{melli}
a greedy lazy model--based
approach for classification was developed
in which the result was a rule tailored 
to the specific observation.
While such an approach gives a simple rule as an answer 
(which is often much easier to understand than a complex rules set)
and often works faster for classification of a single event,
it--as every greedy algorithm--is not guaranteed to find 
the best rule, because the algorithm may not reach
the global maximum of the quality criterion
and a sub--optimal rule may be returned.

In the work \cite{segal} an approach based on the brute force
of rule--space scanning was developed. It was used
for finding the ``nuggets'' of knowledge in the data
(each nugget is a rule with a high degree of correctness).
In contrast with greedy type algorithms,  
massive search algorithms are guaranteed to find
the best rule(s).

In our early work \cite{with_gorodetsky} 
we presented an approach which combined the 
massive model--based rule search approach 
with lazy instance--based learning.
In that work we were also interested in ``nuggets'' of knowledge,
but only those which
were applicable for the instance we wanted to classify.
The result was a set of rules which were applicable 
for classification of the given event.
One may think about these rules as 
a projection of a global classification rules set
to the given instance of the event.

In the current paper this approach is taken to the next 
level, and a practical algorithm, applicable to a variety of
problems, is presented.
A number of significant improvements have been made
since that early version. The current algorithm includes 
the following new features:
1. highly optimized rule--space scanning,
which allows problems with significant number of attributes to be solved;
2. integration of levels selection procedure
for ordered (continuous and literal) attributes with the 
rule search algorithm; 3. information about 
dependent attributes directly included into the 
tree search algorithm thus significantly reducing 
computational complexity; and 
4. an original conflicting rules resolution strategy 
which was especially built to work with automatically 
generated rules.

To create a practical algorithm,
the three aspects --- logical, statistical and computational complexity
need to be addressed.
In section \ref{logical_formulas}
we formulate the problem and discuss the logical formulas 
which represent the rules we are interested in finding.
In section \ref{stat_crit} we discuss 
the statistical quality criterion which can be used 
for evaluation of rule quality and specify the 
criteria which we use in this work.
We also present a conflicting rules resolution strategy
for automatically generated rules.
At the end of section \ref{stat_crit}
a sketch of the algorithm is presented.
In section \ref{var_sel} we discuss the selection 
of attributes for analysis; it should be 
stressed that some attributes as they are built in 
section \ref{var_sel} are not independent, and this fact is known in advance.
In section \ref{complexity} we discuss 
computational complexity issues; an approach which includes
information about dependence of the attributes into the 
algorithm is proposed.
In section \ref{errorestimation} we discuss
error estimation.
In section \ref{data_result} we present the data analysis results 
and compare our results with the results of C4.5R8 \cite{c45R8}.
In section \ref{discussion} a discussion is presented.

\section{Logical formulas as a result of statistical analysis}
\label{logical_formulas}
In this section we describe 
logical formulas obtained 
as a result of data analysis.
Representation of knowledge after it has been learned from the data,
can vary depending on the approach used.
However, different forms of knowledge
representation(decision tables, decision trees, rules list, etc.)
are equivalent to some logical formulas.
Formulas obtained during data
analysis are usually quite complex when 
applied to prediction or classification.
This complicates the understanding of the results. 
The major source of complexity is the fact that 
the formulas are usually built to be applicable to 
all data observations. As we show below
the complexity of the rules can be significantly
reduced if, instead of building global rules,
we build local rules which are defined on a 
subset of observation data; this subset must include
the data point where we want to perform a prediction/classification.
Such an approach combines the best of 
both instance--based and model--based learning.
It can be described as an approach working
with projections of global formulas to local observations.
A drawback of such an approach is the need to recalculate
the rules for every event we want to classify.
This is the cost of using simple local rules instead of 
complex global rules.

In the simplest form the problem 
can be represented as the following:
We have a random variable $g$ (consequent)
and a random vector ${\bf x}$ (antecedent)
of $M$ components $x^{(m)}, m=1\dots M$.
Random variables $g$ and $x^{(m)}$
are assumed to take two different values: $true$ and $false$ (Note that 
this does not limit us in using other types of input data.
The detailed process of $g$ and ${\bf x}$ selection will be described 
in section \ref{var_sel}.)
We have a finite number of observations $N+1$,
each observation gives specific values of
$x^{(m)}_n$ and $g_n$. Index $n=0\dots N$ 
numerates the observations. The value of 
antecedent $x^{(m)}_n$ is known for $n=0\dots N$,
the value of consequent $g_n$ is known for $n=1\dots N$,
at the point $n=0$ the value of consequent is unknown.
The problem is predicting the value of $g$ at $n=0$.
Again,  we are interested in finding a prediction 
of $g$ only at one point $n=0$, 
not in building a universal prediction formula 
which is applicable at any $n$. This 
allows us to build a prediction which
is  easier to build, understand, and interpret.

The prediction is represented as a set of
conjunctive forms
which are correct with a high degree 
of confidence; this set of conjunctives may be considered
as a distinctive conjunctive form of a logical formula.
The criterion of 
acceptance/rejection will be described in section \ref{stat_crit}.
Consider all possible expressions of the form:

\begin{equation}
f=\prod\limits_{m\in\{\mu\}} x^{(m)}_{n}=x^{(m)}_{0} 
\label{xf}
\end{equation}

In the Eq. (\ref{xf}) 
each term is a match of $x^{(m)}$ antecedent component 
at a given point $n$, with the value of the $x^{(m)}$
at the point we want to make a prediction: $n=0$;
index $m$ belongs to a given set of indexes $\{\mu\}$;
we have logical ``and'' in between all these terms,
i.e. the formula (\ref{xf}) represents 
a fact of simultaneous matches of several antecedent 
components (those with indexes in $\{\mu\}$ set)
with their values at the prediction point $n=0$.
Each formula of (\ref{xf}) type is 
completely defined by a set $\{\mu\}$.
In total there are $2^M$ possible 
$\{\mu\}$ sets.

The goal is to find the conjunctive of form (\ref{xf}) which can give an implication 
with high degree of confidence.
\begin{equation}
\left(
\prod\limits_{m\in\{\mu\}} x^{(m)}=x^{(m)}_{0} 
\right)
\rightarrow g=g_0^{(pr)}
\label{implication}
\end{equation}

\noindent 
Formula (\ref{implication}) represents an 
implication rule when a
simultaneous match of given antecedent components 
(those in $\{\mu\}$ set), 
with their values at a point to predict $n=0$,
gives specific values of consequent.
The value  $g_0^{(pr)}$ 
is the value that the rule (\ref{implication}) predicts.
Note that the rule of form (\ref{implication})
is defined 
on a subset of all available observations
(on observations on which (\ref{xf}) is true).
We do not consider the rules
(even if they have very high confidence)
which can not be applied at $n=0$. 
This drastically reduces the number of rules we may accept.

In the next section we discuss 
statistical criteria used for the evaluation of each rule
quality and for resolving the problem of conflicting rules(when several 
high confidence rules predict different values of $g$).

\section{Local prediction rules: statistical evaluation of quality 
	 and conflicts resolution}
\label{stat_crit}

Quality evaluation of 
a rule is based upon its statistical characteristics.
In this paper we use canonical 
statistics: statistics which can be 
expressed via components of matrix of joint 
distribution $(f,g)$:
\begin{equation}
\left(
\begin{array}{lll}
P(f=false,g=false)&;&P(f=false,g=true) \\
P(f=true,g=false)&;&P(f=true,g=true)
\end{array}
\right)
\label{distrib}
\end{equation}

\noindent
Here $g$ is the consequent and $f$ is a logical formula;
for example, one from Eq. (\ref{xf});
the (\ref{distrib}) is 2x2 matrix
(because
both $f$ and $g$ take two different values).
Probability $P$ can be defined in a number of different 
ways. In this paper the probability 
is defined in a standard combinatoric
way (the number of favorite outcomes divided 
by the total number of outcomes).
Almost any of the commonly used (coverage,correctness) type
of criteria can be expressed 
via the components of a matrix (\ref{distrib}).

There are many different statistics
which can be used for quality evaluation of 
a logical formula.
In the work \cite{hh} an approach of logical formulas
transformation was developed which 
may solve an exponential complexity 
combinatoric problem in polynomial time.
A similar approach was developed in \cite{Lyashenko},
where only statistics allowing
formula transformations increasing quality criterion
were used.

A statistic commonly used as a quality criterion is
information gain \cite{shennon}.
The information gain based criterion was used
in a number of machine learning studies.
This criterion usually works well for the evaluation of global rules,
but much less effectively for local rules.
In the case of local rules the major problem 
with information gain criterion 
is the fact that $f\to g$ and $\neg f \to \neg g$
are equally important for this criterion.
For the rules of (\ref{implication}) type
we know in advance that $f=true$ and this asymmetry should be 
included into the quality criterion.
Information gain criterion
work well in the case of a seldom event.
For example, for an event which happens 
in 1 out of 100 cases a rule which predicts
that the event will never happen has $0.99$
correctness. At the same time, an information gain
based criterion gives no value to such a rule
because we get no extra information 
beyond what we already know.

The most widely used statistics
for estimation of a logical formula quality
are ones of (coverage, correctness) type;
the coverage is defined as $P(f=true;g=g_0^{(pr)})/P(g=g_0^{(pr)})$,
and the correctness is defined as $P(f=true;g=g_0^{(pr)})/P(f=true)$.
In \cite{segal} a criterion
based on high correctness (the coverage 
considered to be secondary) has been used.
A criterion based on the F--measure (which combines
precision and recall into one number)
from information retrieval theory \cite{inform_retr}
can also be used as a quality criterion. An important characteristic
of the F--measure is the presence
of a parameter allowing the adjustment of relative
importance of coverage and correctness.

In this paper we use a quality criterion which has 
properties similar to one of (coverage,correctness) type.
The quality $\alpha$ of implication rule is defined as following:
\begin{eqnarray}
\alpha&=&
\lambda \frac{P(f\ne true;g\ne g_0^{(pr)})}{P( g\ne g_0^{(pr)})}+
(1-\lambda) \frac{P(f=true;g=g_0^{(pr)})}{P(g=g_0^{(pr)})}
\label{quality_crit} \\
g_0^{(pr)} &:& P(f=true;g=g_0^{(pr)}) {\rm \ is \ maximal} 
\label{g_max}
\end{eqnarray}

\noindent
In this paper we focus on predicting the events,
not the probabilities, so for a given $f$ we first 
select the value (of two possible values) of $g_0^{(pr)}$ 
which gives maximum of $P(f=true;g=g_0^{(pr)})$,
Eq. (\ref{g_max}), 
then evaluate the quality of implication rule using quality 
criterion (\ref{quality_crit}).
The value of $\alpha$ is equal to $1$ for implications
(\ref{implication}), giving totally correct 
predictions for every observation. For implications
with non--perfect correctness and/or coverage
the value of $\alpha$ is lower than $1$.
The parameter $0\le\lambda\le 1$
determines the relative importance of coverage 
and correctness. The value $\lambda=0.5$
makes coverage and correctness equally important characteristics
of a rule. The values $\lambda>0.5$ make
correctness more important than coverage.

While different statistics give 
very similar results on data which does 
not produce conflicting rules, 
the difference between different statistics
may become  significant when analyzing 
data producing conflicting rules.
Our experiments with different types of data 
have shown that quality criterion (\ref{quality_crit})
works well for the different data that we tested.
To resolve a problem of conflicting rules
we separate the process of making a prediction on two steps.
On the first step
we do not predict the specific value of $g$,
we just find all implication rules of high quality.
On the second step we use all found implication rules
to obtain a prediction.
Let us assume we found all rules of high 
enough quality; for example, with
a quality better than a given acceptance level $\alpha_0$.
Each rule predicts its own $g_0^{(pr)}$ at $n=0$.
If we have no conflicting rules 
(all accepted rules predict the same value of $g_0^{(pr)}$)
everything is very simple: this value is the value we 
predict at $n=0$.
If we have conflicting rules
(rules which predict different values of $g$),
the situation is more complicated, and 
a conflict resolution strategy must be developed.
This is a special problem which 
has been considered in a number of publications.
(See Refs. \cite{brnstwn,durkin,lvdg} for review.)
Most studies focus on resolving
conflicts between hand--crafted, rather than automatically
generated rules. The conflict resolution of 
automatically generated rules has its own specifics.
The simplest approach is to accept only one (the best)
rule. The problem is the fact that 
it is common to have a number of 
rules of similar quality, and the idea of 
taking a single rule and leaving
a number of rules of similar quality 
out of consideration
often causes a significant bias in data analysis.
An approach often used to resolve such conflicts 
is the idea of ordering rules,
but it gives away 
an extremely useful property of rules--based predictions ---
the ability to evaluate rules in arbitrary order.

The approach we use in this paper differs from 
the ones mentioned above in a very significant way.
We assume that all accepted rules must be incorporated 
into the prediction formula. If we do not have conflicting rules
prediction quality usually increases by combining all rules. 
If we do have conflicting rules, prediction 
quality may decrease (often in a very significant way)
when the rules are combined.

For resolving the problem of conflicting rules
consider the following problem:
Let $s$ be a set of observations 
on which the value of $f$ from (\ref{xf}) is true.
The $P(s)$ is the probability of an observation
to give true value of $f$ and $P(g=g_0^{(pr)}\Big/s)$
is the probability of an observation to have $g$ equal to $g_0^{(pr)}$
under the condition that the observation belongs to $s$.
Note that these two probabilities are just equal 
to $P(f=true)$ and $P(g=g_0^{(pr)}\Big/f=true)$ respectively,
but for conflict resolution it is 
much more convenient to work with a set of observations
than with individual rules.
The problem of resolving conflicting rules
is equivalent to the following:
For a number of sets $s_q$, $q=1\dots Q$
determines probabilities of different outcomes of $g$
under the condition that all $s_q$ are true.
For a single rule ($Q=1$) the answer is trivial:
this is either $P(g=g_0^{(pr)}\Big/s)$ or $P(g=g_0^{(pr)})$
depending on whether we accepted or rejected a rule.
For more than one rule ($Q>1$), a formal answer 
can be also written:
 this is either $P(g=g_0^{(pr)}\Big/s_1\cap s_2 \cap \dots \cap s_Q)$
or $P(g=g_0^{(pr)})$ depending on whether we accepted the rules or not.
The problem is that the probability
$P(g=g_0^{(pr)}\Big/s_1\cap s_2 \cap \dots \cap s_Q)$
cannot even be estimated because 
the set $s_1\cap s_2 \cap \dots \cap s_Q$ 
often has few observations, insufficient
for probability calculation.
There is an example of this:
Assume we have 100 observations of $g$ and 
101 observations of $x^{(m)}, m=1\dots 2$,
in the point to predict antecedent ${\bf x}=(true,true)$.
Let $g$ take the value of $true$ on 50 observations 
and $false$ on the other 50.
Suppose we have two implication rules 
$(x^{(1)}=true) \rightarrow (g=false)$ and 
$(x^{(2)}=true) \rightarrow (g=true)$;
both give perfect prediction (correctness and coverage 
are equal to 1) on these 100 observations. 
What will be the probability of different values of $g$ 
in the point to predict ${\bf x}=(true,true)$?
We have two perfect rules. The first one 
predicts $g=false$, and, the second one predicts $g=true$.
The probability $P(g=g_0^{(pr)}\Big/s(x^{(1)}=true)\cap s(x^{(2)}=true))$
cannot be calculated because we have no observation
with known $g$
when $x^{(1)}=true$ and $x^{(2)}=true$ simultaneously.

To resolve such conflicts 
we build a set $S$ from all $s_q$ sets
and then apply a quality criterion
to a single ``combined'' rule 
which is defined on $S$.
This way the problem of conflicting rules
is resolved by introducing a new, ``combined'' rule,
and the answer is the same as the one mentioned above 
for a single rule:
The probability is either 
$P(g=g_0^{(pr)}\Big/S)$ or $P(g=g_0^{(pr)})$
depending on whether we accepted or rejected 
a combined rule.
Having only
one rule we may use a number of different
criteria to evaluate this ``combined'' rule quality;
for example, in addition to criterion (\ref{quality_crit})
we may use $\chi^2$ criterion or any other criteria.
Different criteria usually give similar results
in the case of a single rule
(because adjustment of acceptance level does
not affect how many rules will be accepted/rejected: 
we have only one rule to consider).
It should be stressed here that
the quality of combined rule may be
lower than individual rule quality.
If this happens this often
indicates the presence of rule 
conflicts or data overfitting.

The only problem left to discuss 
is how to obtain the set $S$ from individual 
sets $s_q$. There is no universal way
to do this, because the sets $s_q$ 
are sensitive to the quality criteria.

The simplest way is to choose the set $S$ as
a union of all $s_q$
\begin{equation}
S=s_1\cup s_2 \cup \dots \cup s_Q
\label{S_def}
\end{equation}

\noindent
Returning to a simple example above with 
two perfect conflicting rules: 
the set $S=s(x^{(1)}=true)\cup s(x^{(2)}=true))$
covers all 100 observations
and the criterion (\ref{quality_crit}) produces $0.5$
value ($0.5$ correctness with $1.0$ coverage),
which is a very low value.
The ``combined'' rule must be rejected and 
unconditional probabilities $P(g=g_0^{(pr)})$ should be used
for prediction.
This is what we intuitively expect in such an extreme 
case of conflicting rules.
There are several other ways 
to select the set of observations $S$. 
We will not discuss all the variants here.
The way to select $S$ in (\ref{S_def}) form
seems to work the best for the quality criterion (\ref{quality_crit}).
In addition to that the (\ref{S_def}) way to select $S$
is well protected 
against data overfitting, because overfitted 
rules often produce different values of $g$ 
which drastically reduce combined rule quality. 

Let us return to the original problem
we formulated in the beginning of
section \ref{logical_formulas}.
Now we can present an algorithm 
for predicting the value of consequent at $n=0$.

\begin{enumerate}
\item
Select acceptance level $\alpha_0$.

\item
Initialize set $S$ to an empty set.

\item
 For every set of antecedent indexes $\{\mu\}$ 
 (totally there are $2^M$ sets) do:
   \\ a) Build implication (\ref{implication})
          and evaluate quality $\alpha$ of it.
   \\ b) If $\alpha>\alpha_0$ add all observation
	 points for which $f$ from Eq. (\ref{xf}) is true
	 to the set $S$.
\item
  Evaluate the quality of a ``combined'' rule: 
  the rule which is defined on observations from $S$.
  This can be done by using the same criterion (\ref{quality_crit}),
  $\chi^2$ or any other type of criterion.
  If the combined rule is accepted 
  use $P(g=g_0^{(pr)}\Big/S)$ , if rejected use $P(g=g_0^{(pr)})$ 
  probability to predict the fact of $g$ taking value $g_0^{(pr)}$ at $n=0$.
  The predicted value $g_0^{(pr)}$ 
  corresponds to the event with maximal probability.
\end{enumerate}

\noindent
The algorithm described above 
is of exponential complexity (one needs to check
$2^M$ possible implication rules).
As we will show in section \ref{complexity}
the complexity may be significantly reduced
in an average case.
Before we start discussing 
computational complexity let us discuss
the procedure of attributes selection
for antecedent and consequent.

\section{Selection of attributes for analysis}
\label{var_sel} 
In all of the considerations above,
we always assumed that consequent $g$
and antecedent components $x^{(m)}$ 
are Boolean attributes.
There are
many cases in which the data contain attributes of other types.
In addition to Boolean variables in this paper we consider
continuous variables (variables taking values from an interval)
and discrete (literal) variables (variables taking values
from a finite set of possible values).
The requirement of ordering (so we can compare 
the values which the variable takes)
is very important for analysis, because
this allows us to build an
effective algorithm of levels selection.
The case with non--ordered values 
is much less interesting, because in this case
for a descrete variable
the algorithm described above will use
the following Boolean attribute:
whether the value of the attribute is equal to its value at $n=0$ or not.

Let us consider 
a variable  (continuous or discrete) $r_n$ (index $n=0\dots N$
enumerates the observations)
taking values from some ordered set (for example an interval).
We convert $r_n$ 
to a number of Boolean attributes which will be used as the components
of vector $\bf x$. This transformation is performed
by selecting a grid $y_l, l=1\dots L$ and 
comparing the value of $r$ with levels $y_l$,
that gives antecedent components
$x^{(m(l))}_n=r_n\le y_l$.
The question is how to select levels $y_l$ to use in implication.
The most commonly used approach is 
to take a single level. People usually do this 
because an increase in the number of levels
increases the number of antecedent components
that can drastically increase computational complexity.
The most common criterion used for selection of the split level
is information gain criterion.
In several works \cite{dougherty,Qcont}
this criterion was successfully applied for determination
of levels of comparison.

We propose a new approach for antecedent attributes selection.
The major new characteristics of proposed approach is 
integration of two usually independent steps 
into one step, so the inference algorithm described 
in section \ref{stat_crit}
will perform not only data analysis, but will also 
select levels to compare.

We do not limit ourselves
to one or two levels that we can compare with;
we use a number of levels (the value of $L$ can be chosen
pretty high) and determine 
the real levels to use directly during data analysis.
One may think about this 
as automatic selection of levels in singleton Mamdani rules
in fuzzy logic, see \cite{smr_fuzzy}.

The first step 
is to take an
ordered ($y_l<y_{l+1}$) 
grid $y_l, l=1\dots L$,  which has many different levels.
(The levels $y_l$, may be selected as
all possible values of $r$ or by using 
any of 
supervised or unsupervised discretization techniques \cite{dougherty}.
These levels are only ``initial'' levels.
The inference algorithm will select from these
the ``real'' levels which will be used in implication rules.)
Then we obtain $L$ antecedent components $x^{(l)}=r\le y_l$.
The attributes  $x^{(l)}$ 
are not independent. From the fact of ordering of $y_l$
follows that if $x^{(l)}_n$ is true then $x^{(p)}_n$
is also true for $p>l$. Also
if $x^{(l)}_n$ is false then  $x^{(p)}_n$
is also false for $p<l$.

The second step is to find the highest index $l$ for which $r_0<y_l$ 
is false ($r_0$ is the value of $r$ in the point to predict
$n=0$), and mark this index as $h$.
Then $y_l$ with $l=h,h-1,h-2,\dots , 1$ 
may be considered as lower boundaries of $r$
and $y_l$ while $l=h+1,h+2,\dots , L$ 
may be considered as upper boundaries of $r$.
These upper and lower boundaries 
can be considered as fuzzy levels for $r$.
Instead of determining specific values for
upper/lower levels from 
some {\it ad hoc} special procedure,
we select them during data analysis by using the inference 
algorithm we described in the previous section.
Such integration allows us to automatically select 
the best level for a rule.
While it may look like we have increased the 
number of antecedent components and 
exponentially increased computational 
complexity, this is not really the case.
The difference between standard approach \cite{WittenAndFrank} p.246,
when a $k$--valued
variable is replaced by $k-1$ synthetic Boolean
variables, and our approach is that we incorporate the 
knowledge about the dependence of these  $k-1$ variables 
into the inference algorithm,
In section \ref{complexity} we show that 
this knowledge can drastically reduce 
computational complexity in average case.

The problem of consequent variable selection 
is usually more straightforward than that for antecedents.
If consequent $j$ is a Boolean (literal variable with two values)
nothing special should be done about consequent selection
and we use $j$ as consequent $g$.
If $j$ is an ordered (continuous or discrete) variable then we take a grid 
$y_l, l=1\dots L$, $y_{l+1}>y_l$  and just run the analysis for 
every $g=(j<y_l)$. 
Additional testing on monotonic increase 
of the predicted probability of $true$ value of $g$ with 
increase of $l$ may be performed to test the
consistency of the predictor.
The algorithm in section \ref{stat_crit} 
can also be applied to $g$ taking more than two values,
because the quality criterion (\ref{quality_crit})
may be generalized to such $g$.

\section{Estimation of computational complexity
	 for brute force rules analysis}
\label{complexity}

As we showed in section \ref{stat_crit},
a brute force algorithm is of 
exponential complexity
(it requires $2^M$ rules evaluation).
However, the  
implication rules we consider 
are not independent.
It is often possible to determine from one rule's 
characteristics  
that a set of rules does not have 
a member of required quality,
so that set of rules can be taken out of consideration.
The requirement of preventing data 
overfitting also helps because it eliminates rules that are too complex.
In addition some optimization techniques 
can be applied.
This way we can often 
perform brute force analysis for a 
problem with a significant number of components.

Let us discuss the properties which allow us
to reduce computational complexity.
\begin{enumerate}
\item
\label{max_rule_complexity}
Preventing data overfitting. 
This usually requires taking out of consideration
overly complex rules.
We do this by considering only rules 
with less that $M_{max}$ terms ($M_{max}<M$) 
in implication (\ref{implication}).
This immediately reduces the number of rules 
to consider from $2^M$ to 
$C_M^0 +C_M^1 + \dots +C_M^{M_{max}}\approx \frac{M^{M_{max}}}{M_{max}!}$
which is still too high.

\item 
Taking into account dependent antecedent components.
In this paper we consider the simplest case: upper and lower 
boundary  
antecedent variables as we build them in 
section \ref{var_sel}.

Antecedent attributes as we build them 
in section \ref{var_sel} from variable $r$ 
(which takes values in some ordered set)
 are not independent.
For example the components  of lower boundary
$r_n<y_l$ with $l=h,h-1,h-2,\dots , 1$
have the following property
\begin{eqnarray}
&&\left(x_n^{(m(l_1))}=x_0^{(m(l_1))}\right)
\&
 \left(x_n^{(m(l_2))}=x_0^{(m(l_2))}\right)= \nonumber \\
&& \phantom{x_n^{(m(l_3))}}
  \left(x_n^{(m(l_3))}=x_0^{(m(l_3))}\right) \label{l_dependent} \\
&& l_3=\max(l_1,l_2) \noindent 
\end{eqnarray}

\noindent
The property (\ref{l_dependent}) 
follows from the fact of 
ordering of $y_l$, the way of $h$ selection
which leads to $x_0^{(m(l_1))}=x_0^{(m(l_2))}=false$
and the following equation:
\begin{equation}
(r<a)\&(r<b)=(r<\min(a,b))
\label{c_dependent}
\end{equation}

\noindent
An equation very similar to (\ref{l_dependent})
can be also written for upper boundary set
$r_n<y_l$ with $l=h+1,h+2,\dots , L$.
This means that only one attribute 
from the upper(lower) boundary set
needs to be included in implication (\ref{implication}).
If we put two components from the upper(lower) boundary
set of attributes then, by applying a (\ref{c_dependent}) type
of transformation, we can always replace two terms 
by a single one.
This property, which is known directly
from the antecedent, allows us to reduce 
the number of implications we need to consider.
Increase in computational complexity
when adding one set with $n_d$ dependent antecedent components
selected as described above in terms of computational 
complexity is equivalent to adding 
much fewer  (about $\log_2(n_d+1)$) independent 
components.
This is why we can integrate selection of fuzzy levels 
with the inference algorithm without much 
increase in computational complexity.
Addition of $L$ levels to test is equivalent 
to adding about $\log_2(h+1)+\log_2(L-h+1)$
independent Boolean attributes.

\item
If an implication rule of (\ref{implication}) form
has a perfect (or close to perfect) correctness,
then the quality of this rule
can not be improved by adding more 
elements to set $\{\mu\}$ (see \cite{segal}),
because by adding more conditions 
we just decrease coverage 
while correctness cannot be further improved.
This means we do not need to consider
the subsets of rules with close to perfect correctness.

\item
\label{mincoverage}
As it has been shown in \cite{segal},
an implication rule (and all rules which include it) 
with coverage below some level 
cannot produce a rule of the required quality.
This requirement can be slightly improved 
by using minimal probability $p^{(v)}$
for every consequent value (the $v\in\{true,false\}$ 
is one of two possible consequent values).
Specifically, for at least one $v$ we must have
$P((g=v)\&(f=true))>p^{(v)}$.
If we have no single $v$ for which this condition
holds, then the implication (and all rules which include it)
cannot produce a rule of the required quality.
For the quality criterion (\ref{quality_crit}) 
the value of $p^{(v)}$ can be easily obtained
\begin{eqnarray}
p^{(v)}&=&\frac{(\alpha_0-\lambda)P(g=v)}{1-\lambda}
\label{minMatches} 
\end{eqnarray}

\item
\label{redundand}
An implication rule must not have 
redundant conditions.
An extreme example of redundant condition 
is a situation when a term $x_n^{(m(l))}=x_0^{(m(l))}$
is added to implication (\ref{implication}) twice.
This does not change any property of a rule,
it just increases the complexity of it.
To check for redundancy of a rule with $m$
conjunctions we may compare the rule
with $m$ rules obtained by taking out one 
condition from the original rule,
see section 7.3.13 (page 318), Ref. \cite{hh}.
Specifically in our case this criterion can be formulated 
as following:
Having a $\{\mu\}$ set with $m$ elements
consider $m$ formulas $f_m$ of (\ref{xf}) type,
each one is obtained by taking out one of $m$ element.
If for at least one $f_m$ 
there is no $v$ for which the 
condition
$P((f_m=true) \& (f=false) \& (g=v))>p_{mism}^{(v)}$
holds, then 
this rule (and all rules which include it)
have redundant conditions and should not be considered.
The value of $p_{mism}^{(v)}$ can be obtained 
from the same formula (\ref{minMatches}) which was used
for $p^{(v)}$. The only difference is the different value 
of threshold $\alpha_0$. For mismatches, the 
threshold $\alpha_1$ is usually chosen lower that $\alpha_0$.
\end{enumerate}

The five properties presented above allow us to
build an algorithm of polynomial
complexity.
This comes from the fact that we are interested 
only in rules applicable at $n=0$, 
what reduces the number of rules to consider 
from $2^{2^M}$ to $2^M$ and from
the properties \ref{mincoverage} and \ref{redundand}
which limit the maximal tree depth in a typical case.
In the worst case the tree depth
is limited by the value of $M_{max}$ from item \ref{max_rule_complexity}.
The other properties reduce the complexity further.
This algorithm, which in typical case is of polinomial complexity 
on $N$ and $M$, can be applied for
solving a variety of practical problems. 

The algorithm 
can be applied to
a brute force analysis for 
a problem with a significant number of components.
A sketch for the algorithm is the following:
All possible implication rules may be 
represented as a tree.
Each node has an antecedent index assigned to it.
Every node can be mapped to a $\{\mu\}$ set
(by taking indexes of this node and all its ancestors).
This property means that if node $A$ is an ancestor of node $B$,
then $f_B=f_A \& X$ where $f_A$ and $f_B$ are 
formulas of (\ref{xf}) type obtained from a $\{\mu\}$ 
set corresponding to nodes $A$ and $B$ respectively,
that allows us to implement the algorithm
as a recursive tree scanning algorithm 
and directly incorporate five properties 
above as indicators for a branch not
having a rule of the required quality.
We discuss different applications in section \ref{data_result}.

\section{Predictor: error estimation}
\label{errorestimation}

An estimation of predictor correctness 
usually involves
building a global rule on training data 
and then evaluating this rule's quality on testing data. 
While this testing approach suits well for 
testing global rules, it is not very convenient 
when considering local rules,
because for every prediction point we may 
have different local rules.
It is nice to know the quality of a local rule,
but this information is not useful
for error estimation at the other prediction points.

The best way to perform testing in such a case is to test
the average performance of the predictor.
One may consider a predictor as some kind of ``global rule''
and estimate its quality. The quality of such a ``global rule''
is equivalent to the predictor average quality.

A common problem of errors estimation
is the limited number of observations.
Techniques such as bootstrap and cross--validation 
are commonly used for performing error estimation 
with a limited number of observations.

For local predictors, a leave--one--out type of cross--validation
is very promising when working with a
limited number of observations. 
This type of testing includes
creation of a set with $N-1$ observations and this
data is used for predicting the value at one left point with
known value of $g$. The procedure is repeated
$N$ times and average predictor performance is obtained.
Mentioned in \cite{WittenAndFrank}  the non--stratification 
problem of testing data (the data in every testing set
has only one observation and
does not have the right proportions of observations 
with different values of $g$)
is much less an issue in the case of local 
predictions than in the case of global predictions,
because the predictor was specifically built to be applicable 
at the point where it tested.

In case we have plenty of data, 
we can estimate predictor average performance 
without leave--one--out cross--validation.
The fact of the local nature of the 
predictor should be taken into account when performing 
the tests. 
Assume we have a training set of $N$ observations
and testing set of $T$ observations.
To determine predictor average performance 
we predict the value for every observation
in a testing set using all the observations
from the training set. In total we run predictor $T$
times (for every observation from testing set) 
each time using the same training set with $N$ observations
and estimate predictor average performance 
from these $T$ predictor runs. 
Predictor average correctness $C$ is defined as:

\begin{eqnarray}
C&=&\sum\limits_{j=true,false} p_{jj} \label{correctness} \\
p_{jk}&=&\frac{t\left((g=j)\&(g^{(predicted)}=k)\right)}{T} \label{pqr} 
\end{eqnarray}

\noindent
The probabilities in (\ref{correctness}) are calculated
in the testing space; 
the value of $t\left((g=j)\&(g^{(pr)}=k)\right)$ 
is the number of tests (totally there are $T$ test runs)
when the value of the consequent which really happened 
was equal to $j$ and the predicted value was $k$.

One of the problems with (\ref{correctness}) and similar 
types of criteria is its dependence
on unconditional probabilities of different outcomes of $g$.
For example, if we have an event which happens
in 1 out of 100 cases, then a predictor 
predicting that the event will never happen
has $0.99$ correctness. This high 
value of correctness is not a result
of predictor quality but of 
the distribution of $g$.
One may use information gain based criteria, 
but using several criteria 
simultaneously complicates the analysis.
This problem does not arise when the (\ref{correctness})
criterion is used for relative comparison 
of different predicting techniques on the same data, 
because in this case we have identical distribution of $g$.

\section{Results of data analysis}
\label{data_result}

The real algorithm has a number of features 
not presented in the basic algorithm 
description which we gave in sections 
\ref{stat_crit}, \ref{var_sel} and \ref{complexity}.
These are some of them:

1. The acceptance level $\alpha_0$ is dynamically adjusted.
	First we set an initial acceptance level.
	Then, during tree scanning, required acceptance 
	level gets automatically increased 
	to $\kappa \alpha$
	if we find a rule with quality $\alpha$ such us
	$\alpha_0<\kappa \alpha$,
	i.e. we keep only rules with quality better 
	than $\kappa$ fraction of the best rule quality;
	all rules with the quality below this value are pruned.
	The value $\kappa=1$ corresponds to the case 
	when only the best rule is accepted.

2.  Dependent variables are also handled 
	in a slightly 
	more complex way than described  
	because of additional optimization.

These and other details which are not described 
here make the algorithm practical.
This algorithm was implemented in
the MLS program (Massive Local Search),
the complete source code of
which is available from \cite{website}.

The following parameters were used during all trials.
The parameter $\lambda$ in quality criterion (\ref{quality_crit})
was set to $0.75$ making correctness more important than coverage.
The maximal tree depth $M_{max}$ 
was set to 8.
The $c_{min}$ was set to $0.08$,
i.e. we accept only rules with quality better than
the quality of a perfectly correct rule covering $0.08$ of positive samples
(with the exception of Chess, Mushroom, Spambase for which $c_{min}=0.17$ was used).
The minimal number of mismatches (item \ref{redundand} 
in section \ref{complexity}) was also determined 
on a base of minimal coverage;
the value of $c_{min}^{(mism)}$ was set to $0.02$
(with the exception of Chess, Mushroom, Spambase for which $c_{min}^{(mism)}=0.1$
was used).
This threshold stayed the same during tree scanning
and was not adjusted as it was for the matches.

For non--ordered input variables antecedent components were 
built as a fact of the exact match of variable value with 
its value at a point to predict.
For ordered input variables (with the exception of Ionosphere and Spambase
for which we used exact match variables)
antecedent components were 
built as described in section \ref{var_sel}.
For ordered literal variables
we used all possible values as initial levels $y_l$, $l=1\dots L$.
For continuous variables a discretization was performed 
first to build ordered literal variables, then 
the same attribute selection procedure used for ordered literal 
variables was applied.
The initial levels for continuous attributes 
may be selected in a number of different ways,
for sufficiently big $L$ different supervised and unsupervised techniques
give very similar results for predictor quality.
 From a computational complexity point of view it is 
good to have low values of $L$.
The entropy based discretization \cite{dougherty}
gives a very good balance of quality and levels numbers.
In this work the entropy based discretization from \cite{dougherty}
was used for initial selection of levels $y_l$ 
for continuous variables, then the procedure 
described in section \ref{var_sel} was applied.
The utility we used is available from \cite{mlc}.

A rich collection of data from
UCI repository \cite{uci}
allows a comprehensive  data
analysis on data
from different domains to be performed. 
Predictor correctness was estimated 
using 3--fold cross--validation with stratification.
Obtained results were compared with ones produced by widely 
used program C4.5R8 \cite{c45R8} with default settings.
In Table \ref{table_res} we present the comparison
of MLS with C4.5R8.
For comprehensive comparison with 
the other predicting techniques
we refer to \cite{lim,zweb,gb},
where a variety of predicting techniques
were tested on the same data from UCI repository.
The error estimation from these works 
can be directly compared 
with ones from Table \ref{table_res} of this paper,
which allows our technique to be easily compared
with the other predicting techniques.

The exceptions mentioned above in algorithm parameter
values for some datasets (Chess, Mushroom and Spambase)
were required to reduce computation time.
The higher values of $c_{min}$ and  $c_{min}^{(mism)}$
the earlier tree scanning algorithm, will reach termination criteria.

The first column of Table \ref{table_res} identifies the data set.
The second and third columns contain 
predictor correctness $C$ for C4.5R8 and our program MLS respectively.
The fourth column contains the total number of observations.
(These are needed for calculation of correctness 
error due to the finite number of tests run.
For a given confidence level and number of tests run
the lower boundary of $C$ can be estimated using 
standard statistical technique \cite{mjser}.
We do not demonstrate this analysis here because we are interested only
in comparison of two predicting techniques.)
The number of antecedent variables
is presented in the last column. This value is for estimation of
computational complexity.
(Note that the number of antecedent
components typically higher than the number of variables 
because the methodology from 
section \ref{var_sel} usually gives several 
antecedent components for a single variable.)

\begin{table}[t] %
\caption[]{MLS and C4.5R8 performance comparison.}
\begin{tabular}{lllrr}
\hline
data            & C4.5R8 & MLS & $N_{observations}$ & $N_{variables}$ \\
\hline
Monk1		&1.0	&1.0	& 432 & 6 \\
Monk2		&0.65	&0.71	& 432 & 6 \\
Monk3		&1.0	&0.972	& 432 & 6 \\
Breast-cancer	&0.73	& 0.69	& 286 & 9 \\
Chess		&0.99	& 0.90	& 3196 & 36 \\
Crx		&0.82	&0.86	& 690 & 15 \\
Diabetes	&0.74	&0.78	& 768 & 8  \\
Hepatitis	&0.75	&0.86	& 155 & 19 \\
Horse-colic	&0.8	&0.81	& 368 & 22 \\
Ionosphere	&0.90	&0.90	& 351 & 34 \\
Labor-neg	&0.72	&0.81	& 57  & 16 \\
Mushroom	&1.0	&0.96	& 8124 & 22 \\
Pima		&0.74	&0.77	& 768 & 8  \\
Spambase	&0.92	&0.87	& 4601 & 57 \\
Tic-tac-toe	&0.985	&0.99	& 958  & 9  \\
Vote		&0.96	&0.96	& 435  & 16 
\end{tabular}
\label{table_res}
\end{table}

The trials are usually executed
much faster in C4.5R8 than MLS.
First, because C4.5R8 is written in C while our program MLS is
written in Java. 
Second, because we need to re--run 
the predictor for every test (lazy learning), 
while C4.5R8 does this only once (eager learning).
This slowdown is important only when doing
predictor testing, because we are especially interested 
in the tasks when just a few predictions,
not about the same number as the training set, 
is necessary.
Third, massive search algorithms are generally 
slower than decision tree ``divide and conquer'' type of algorithms.
Despite running more slowly, the proposed algorithm 
is fast enough to solve practical problems.

The Monk1, Monk2 and Monk3 are
the problems usually tried first by 
different predictor algorithms.
 From Table \ref{table_res} it follows that 
on monk tests
MLS performs about the same 
or slightly better than C4.5R8.

On Chess MLS performs noticeably worse
that C4.5R8. This is because the values of $c_{min}$ and $c_{min}^{(mism)}$
used for this trial effectively reduce maximal tree depth to 
a value of about 4.
At the same time, C4.5R8 generates a number of rules
with more than 10 conditions.
This Chess problem is an example of a
problem for which massive search approach is not effective:
a large number of attributes produce complex rules.
A similar effect (but to much lower degree) occurs
in the Mushroom trial.

On Crx,
Diabetes,
Hepatitis,
Horse-colic,
Labor-neg,
Pima,
Tic-tac-toe
 and 
Vote
MLS performs about the same as or better than C4.5R8.
These trials also have significant number of attributes,
but the rules do not have too many conditions,
and global optimization algorithm easily catches the
best rule(s) 
without any major slowdown in calculations.

Our tests also show that in some trials 
different values of $c_{min}$ (minimal coverage)
and parameter $\lambda=0.75$ from Eq. (\ref{quality_crit})
(relative importance of quality and correctness)
may result in better correctness 
than presented in Table \ref{table_res}.
The higher $c_{min}$ the higher is the required quality 
for a rule to be accepted.
As shown in section \ref{complexity},
the value of $c_{min}$
affects computational complexity. The increase 
of $c_{min}$ and $c_{min}^{(mism)}$
decreases computational complexity by reducing
the effective tree scanning depth.
We expect that automatic adjustment of 
parameter $\lambda$ and required  minimal coverage $c_{min}$
based on available data will make noticeable improvement
to MLS.

In addition to predictor quality on different datasets
another thing we are interested in testing is an effect
of attribute selection methodology from
section \ref{var_sel} to predictor quality.
To test this we ran the predictor twice on some datasets:
the first time all antecedents were selected as a fact of 
exact match of variable value 
with its value at a point to predict,
and the second time all antecedent components
(even if they correspond to non--ordered attributes)
were selected as a
comparison with upper and lower levels in the way
described in section \ref{var_sel}.
Note that the former selection 
can be always obtained from the latter one
because the condition $r=a$ is the same as $(r\le a)\&(r>a-1)$
(here the variable $r$ assumed taking integer values from an interval).
This way we tested how the quality of a predictor
is affected by the increase of rule expressive power
when we go from ``exact match'' type of attributes
to the type of attributes built in the way
described in section \ref{var_sel}.
We performed this testing on five datasets with 
ordered attributes
(in Monk1, Monk2 and Monk3 the structure of attributes 
values 
allows the variables being considered as ordered,
and in Pima and Diabetes the attributes are ordered),
and two datasets with literal non--ordered attributes
(Tic-tac-toe and Vote).
The results are presented in Table \ref{table_emm}

\begin{table}[t] %
\begin{tabular}{lrrrrr}
\hline
data & Exact Match & Levels Comparison \\
\hline \\
Monk1 & 1.0  &  1.0 \\
Monk2 & -  & 0.71  \\
Monk3 & 0.972  & 0.972  \\
Pima  & 0.76   & 0.77   \\
Diabetes & 0.78 &  0.78 \\
Tic-tac-toe & 0.99 & 0.82 \\
Vote        & 0.96 &  0.94  \\
\hline
\end{tabular}
\caption[]{Predictor results for ``exact match'' and ``level comparison'' type of attribute selection for MLS.}
\label{table_emm}
\end{table}

 From these trials it follows that for datasets 
with ordered attributes (Monk1, Monk2 and Monk3)
the transition from ``exact match'' to ``levels comparison''
may significantly increase predictor quality.
In Monk2 no single rule found for an exact match (because we 
require high enough minimal coverage for a rule),
but increased expressive power of generated rules
allows us to generate high quality rules which obey
the condition of minimal coverage.
At the same time in trials where antecedent attributes were obtained 
as a result of entropy discretization (Diabetes and Pima)
there is no strong effect 
of automatic selection of upper/lower boundaries.
Because of computational complexity 
for Ionosphere and Spambase we did only ``exact match'' type of
attribute selection trials, but preliminary results 
show that ``exact match'' type of attributes may 
produce even better results than ``levels comparison''.

For the datasets with non-ordered attributes (Tic-tac-toe and Vote,
for which  we forced non-ordered variables being considered as ordered)
such transition may either not affect or even decrease predictor quality.
This is because increased expressive power of the rules
may cause an effect similar to data overfitting.
The most clear example is Tic-tac-toe trial,
where the global optimization algorithm finds 
many ``false rules'', a combination 
of conditions which by chance happened to give
a high value of quality criterion. Such ``false rules''
can be significantly reduced by increasing the value of minimal coverage.

 From these trials it follows that the approach 
to antecedent attributes selection 
from section \ref{var_sel}
may give better results only for ordered attributes,
and even in this case an
``exact match'' of attributes may produce
better results in some instances.

Presented  test trials show that
the massive search algorithm often performs
about the same or better than C4.5R8 on many datasets.
We attribute this to global optimization.
There are also cases when MLS is less effective than 
decision tree ``divide and conquer'' type of algorithms.
This usually happens on the datasets with a large number of attributes 
producing complex rules.

\section{Discussion}
\label{discussion}
The described approach proves
that a massive local rules search global optimization algorithm
can be applied to problems with a significant
number of attributes.
The computational complexity can be greatly
reduced by building rules which are specific 
to a prediction point and by using the optimization technique
described above. The massive search algorithm
is guaranteed to find the global maximum 
which makes it especially valuable for testing
various predicting systems.

In this work we have shown that 
the process of attributes selection  can be integrated
with the process of rules search. This allows us to perform
data analysis in a uniform way without separation of
the attribute selection and the rules search stages.
 From a fuzzy logic approach this may be considered
as automatic selection of levels in
singleton Mamdani rules \cite{smr_fuzzy}.
Such a method of attribute selection
usually allows us to build more ``expressive'' rules.
This is related to the fact that in many problems
the comparison of the value with a level is 
a natural method of attribute selection for the problem.

Another distinctive feature of the proposed algorithm 
is a conflicting rules resolution strategy.
We accept a number of rules, then
build a single rule for prediction based on accepted rules.
The quality of this single rule may significantly decrease
if accepted rules predict different values of consequence.

While the described approach 
is already practical and was applied in the 
solution of a number of different problems,
it can be further improved. From our point of view
there are two improvements which would improve the algorithm.
Firstly, the quality criterion (\ref{quality_crit}) 
is different than commonly used criteria.
The major advantage of the criterion is the 
fact that its calculation can be optimized.
The problem of probability calculation of a logical expression 
is a problem actively studied from a 
computational complexity point 
of view, see \cite{abraham,heidman,bm,prob_gorodetsky,ba}
and references therein.
Because calculation of (\ref{quality_crit}) is equivalent 
to calculation of a probability various optimizations
used in reliability theory \cite{ln} can be applied.
Another improvement which can be added to the algorithm 
is automatic selection of minimal coverage $c_{min}$ 
and relative importance of coverage and correctness $\lambda$. 
A flexible selection of these parameters
often improves the results.
These improvements, in our opinion,
can further increase the correctness and decrease
the computational complexity of the algorithm.

\acknowledgements
Vladislav Malyshkin greatly appreciates Columbus Advisors LLC's support for this study, especially the
 support from Emilio J. Lamar during Vladislav's employment with Columbus Advisors LLC. 
The authors would also like to thank Alexander Rybalov 
for many fruitful discussions. 

\theendnotes

\end{article}
\end{document}